%% file: acl2019.tex
\documentclass[11pt,a4paper]{article}
\usepackage[usenames, dvipsnames]{xcolor}
\usepackage[hyperref]{acl2019}
\usepackage{stmaryrd}
\usepackage{times}
\usepackage{latexsym}
\usepackage{graphicx}
\usepackage{caption}

\usepackage{amsmath}
\usepackage{tabularx}
\usepackage{mathtools}
\usepackage{booktabs}
\usepackage{url}
\usepackage{longtable}
\usepackage{tabu}
\usepackage{subfigure}
\usepackage{bbm}
\usepackage{paralist}

\usepackage{multirow}

\aclfinalcopy % Uncomment this line for the final submission
\newcommand{\token}[1]{\raisebox{-2pt}{\includegraphics[height=1em]{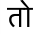}}}
\newcommand{\noun}[1]{\raisebox{-2pt}{\includegraphics[height=1.5em]{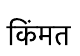}}}
\newcommand{\suffix}[1]{\raisebox{-2pt}{\includegraphics[height=1.2em]{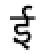}}}

\title{CMU-01 at the SIGMORPHON 2019 Shared Task on \\
Crosslinguality and Context in Morphology}

\author{Aditi Chaudhary \hspace{2mm}
    Elizabeth Salesky \hspace{2mm}
    Gayatri Bhat  \\
   {\bf David R. Mortensen}  \hspace{2mm}
   {\bf Jaime G. Carbonell} \hspace{2mm}
  {\bf Yulia Tsvetkov} \\
  {\tt \{aschaudh, esalesky, gbhat, dmortens, jgc, ytsvetko\}@cs.cmu.edu}\\
  Language Technologies Institute \\
  Carnegie Mellon University
 }

\date{}

\begin{document}
\maketitle
\begin{abstract}
This paper presents the submission by the CMU-01 team to the SIGMORPHON 2019 task 2 of Morphological Analysis and Lemmatization in Context. This task requires us to produce the lemma and morpho-syntactic description of each token in a sequence, for 107 treebanks. We approach this task with a hierarchical neural conditional random field (CRF) model which predicts each coarse-grained feature (eg. POS, Case, etc.) independently. However, most treebanks are under-resourced, thus making it challenging to train deep neural models for them. Hence, we propose a multi-lingual transfer training regime where we transfer from multiple related languages that share similar typology.\footnote{The code is available at \url{https://github.com/Aditi138/MorphologicalAnalysis/}.} 
\end{abstract}

\input{intro.tex}
\input{method.tex}
\input{experiments.tex}

\input{Results.tex}

\input{conclusion.tex}

\bibliography{acl2019}
\bibliographystyle{acl_natbib}

\appendix
\newpage
\section*{Appendix}
\input{AResults.tex}

\input{Amethod.tex}
\section{Analysis}
\label{langanalysis}
\begin{figure*}
    \centering
    \subfigure[Belarusian (be-hse)]{%
    \label{be}%
    \includegraphics[width=\textwidth]{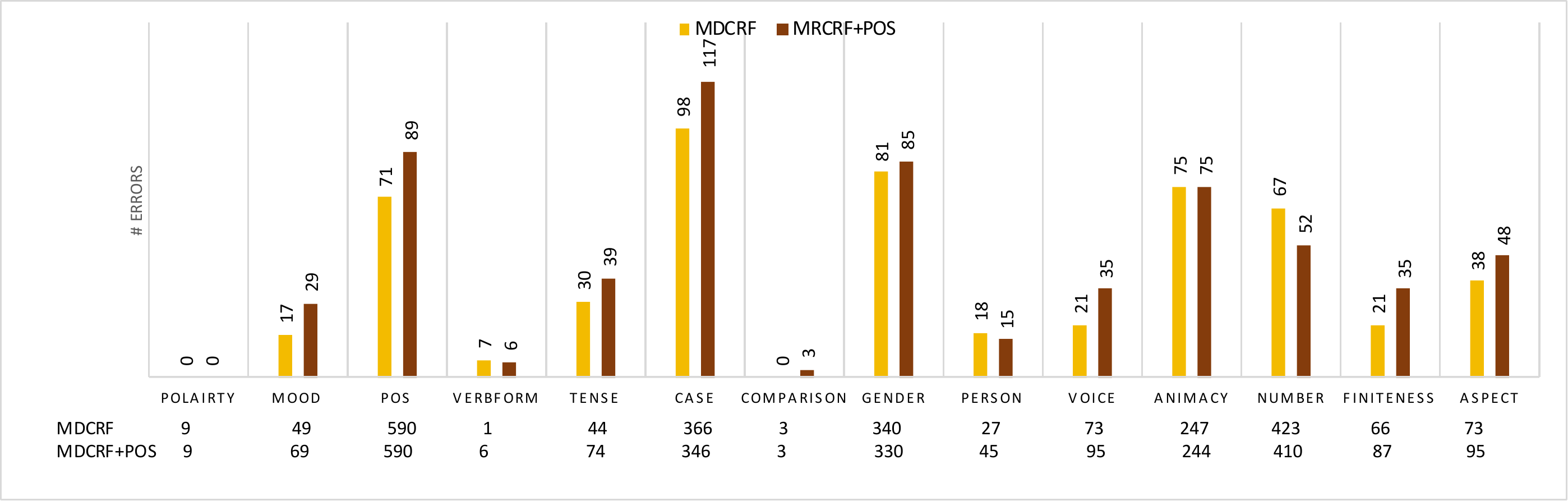}}
    \subfigure[Ukrainian (uk-iu)]{%
    \label{uk}%
    \includegraphics[width=\textwidth]{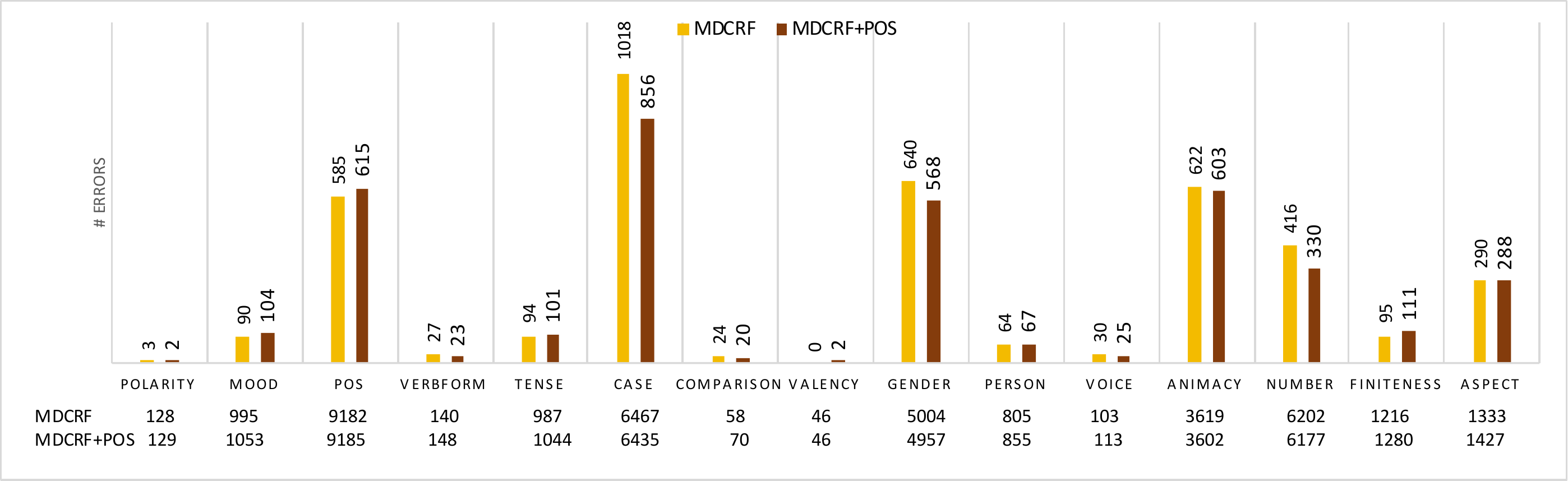}}
    
     \subfigure[Sanskrit (sa-ufal)]{%
    \label{sa}%
    \includegraphics[width=\textwidth]{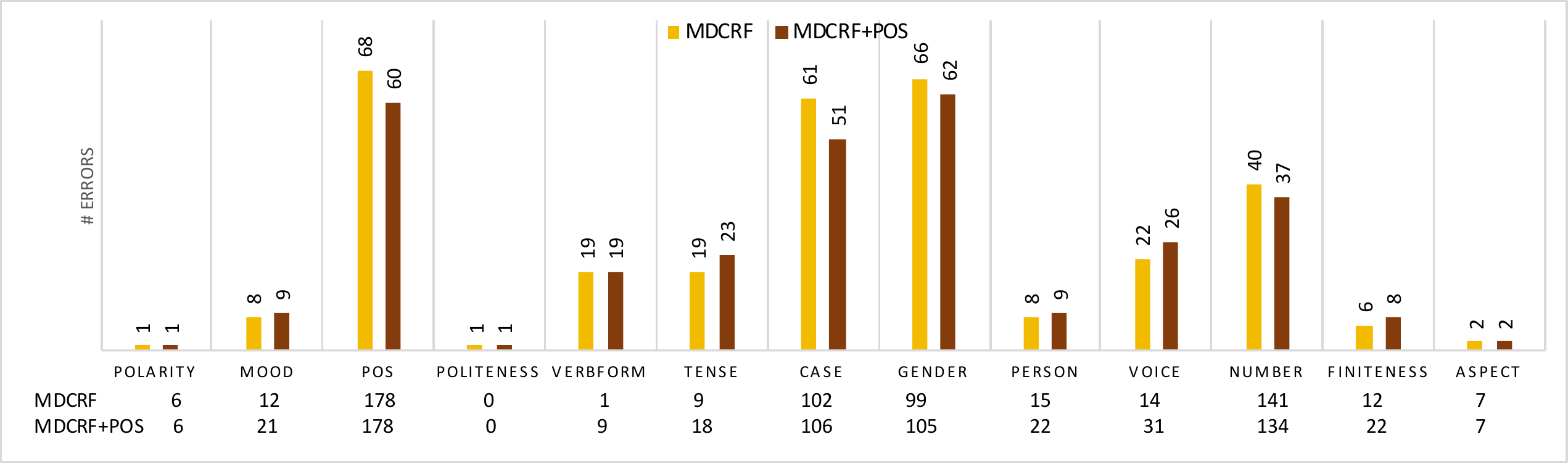}}

  \caption{Number of errors per coarse-grained feature for models comparing the addition of POS to the encoder. The rows at the bottom denote the total number of predictions per each feature for both the models.}%
\label{fig:eg5}%
\end{figure*}

In order to understand where the addition of POS helps, we plot the number of errors per each coarse-grained feature for three languages in Figure \ref{fig:eg5}. For Sanskrit and Ukrainian we see that POS generally helps reduce the errors predominantly for the features: Case, Gender, Number. For Belarusian, we did not observe a clear trend since the POS accuracy actually decreased for \textsc{mdcrf+pos}.

\end{document}

%% file: intro.tex
\section{Introduction}
\label{intro}
Morphological analysis \cite{hajic1998tagging,oflazer1994tagging} is the task of predicting morpho-syntactic properties along with the lemma of each token in a sequence, with several downstream applications including machine translation \cite{vylomova2016word}, named entity recognition \cite{gungor2018improving} and semantic role labeling \cite{strubell-etal-2018-linguistically}. Advances in deep learning have enabled significant progress for the task of morphological tagging \cite{muller-schuetze-2015-robust,heigold-etal-2017-extensive} and lemmatization \cite{malaviya2019simple} under large amounts of annotated data. However, most languages are under-resourced and often exhibit diverse linguistic phenomena, thus making it challenging to generalize existing state-of-the-art models for all languages. 

In order to tackle the issue of data scarcity, recent approaches have coupled deep learning with cross-lingual transfer learning \cite{malaviya2018neural, cotterell2017cross, kondratyuk201975} and have shown promising results. Previous works \cite[e.g.,][]{cotterell2017cross} combine the set of morphological properties into a single monolithic tag and employ multi-sequence classification. This runs the risk of data sparsity and exploding output space for morphologically rich languages. \citet{malaviya2018neural} instead predict each coarse-grained feature, such as part-of-speech (POS) or Case, separately by modeling dependencies between these features and also between the labels across the sequence using a factorial conditional random field (CRF). However, this results in a large number of factors leading to a slower training time (over 24h).

To address the issues of both data sparsity and having a tractable computation time, we propose a hierarchical neural model which predicts each coarse-grained feature independently, but without modeling the pairwise interactions within them. This results in a time-efficient computation (5--6h) and substantially outperforms the baselines.  
To more explicitly incorporate syntactic knowledge, we embed POS information in an encoder which is shared with all feature decoders.
%(note we probably use pos a few sentences up, is it okay to define here?
% To facilitate syntactic knowledge, we incorporate part-of-speech (POS) information in our neural encoder which is shared with all the feature decoders.
To address the issue of data scarcity, we present two multilingual transfer approaches where we train on a group of typologically related languages and find that language-groups with shallower time-depths (i.e., period of time during which languages diverged to become independent) tend to benefit the most from transfer.
We focus on the task of contextual morphological analysis and use the provided baseline model for the task of lemmatization  \cite{malaviya2019simple}. 

This paper makes the following contributions:
\begin{asparaenum}
    \item We present a hierarchical neural model for contextual morphological analysis with a shared encoder and independent decoders for each coarse-grained feature. This provides us with the flexibility to produce any combination of features. 
    \item We analyze the dependencies among different morphological features to inform model choices, and find that adding POS information to the encoder significantly improves prediction accuracy by reducing errors across features, particularly Gender errors. 
    \item We evaluate our proposed approach on 107 treebanks and achieve +14.76 (accuracy) average improvement over the shared task baseline \cite{sigmorphon2019} for morphological analysis. 
\end{asparaenum}

% \section{Task Description}
% \label{taskdes}
% Task 2 of the SIGMORPHON 2019 Shared Task is Morphological Analysis and Lemmatization in Context. Given a sequence, the aim is to generate the lemma and morphological syntactic description (MSD) for each token. This year's edition features a total of 107 treebanks spanning 66 languages, drawn from the Universal Dependencies \cite{nivre2016universal} project. The MSDs have been converted to the UniMorph schema (UM) \cite{kirov-etal-2018-unimorph},for annotation consistency across languages. The consistent annotation scheme enables our proposed multilingual transfer method.

%% file: method.tex
\section{Contextual Morphological Analysis}
\label{method}
In this section, we formally define the task (\S \ref{taskform}) and  describe our proposed approach (\S \ref{ourmethod}).
\subsection{Task Formulation}
\label{taskform}
Formally, we define the task of contextual morphological analysis as a sequence tagging problem.  Given a sequence of tokens $\mathbf{x} = x_1, x_2, \cdots , x_n$, the task is to predict the morphological tagset $\mathbf{y} = y_1, y_2, \cdots, y_n$ where the target label \textit{$y_i$} for a token \textit{$x_i$} constitutes the fine-grained morpho-syntactic traits \{\textit{N;PL;NOM;FEM}\}.

\begin{figure*}
 \centering
  \includegraphics[width=0.8\textwidth]{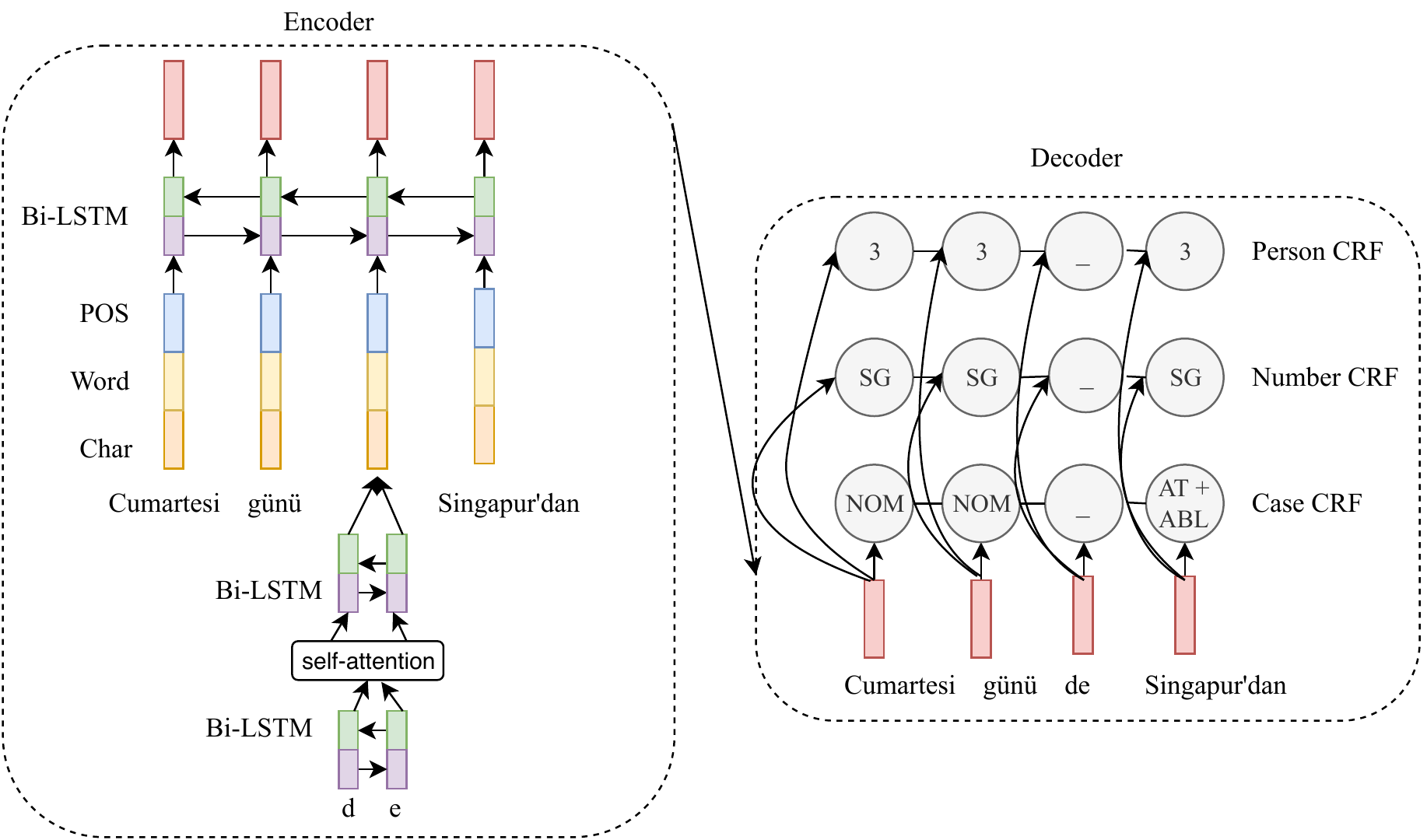} %\textwidth]{images/sigmorphon.pdf}
  \caption{Hierarchical neural model for contextual morphological analysis with independent CRF decoders for each coarse-grained feature $F$. For the model \textsc{mdcrf+pos}, POS embeddings are concatenated to the word and char-level representations as depicted above. This model has $|F|$-1 decoders since POS tagger is run separately as a prior step. \textsc{mdcrf} refers to the above model without POS embeddings having all $|F|$ decoders.  }
  \label{fig:model}
\end{figure*}

\subsection{Our Method}
\label{ourmethod}
In line with \citet{malaviya2018neural}, we formulate morphological analysis as a feature-wise sequence prediction task, where we predict the fine-grained labels (e.g N, NOM, ...) for the corresponding coarse-grained features $F=$\{POS,Case,...\} as shown in Figure~\ref{fig:model}. However, we only model the transition dependencies between the labels of a feature. This is done for two reasons: 1) As per \citet{malaviya2018neural}'s analysis, the removal of pairwise dependencies led to only a -0.93 (avg.) decrease in the F1 score. We further observe in our experiments that our formulation performs better even without explicitly modeling pairwise dependencies; 2) The factorial CRF model gets computationally expensive to train with pairwise dependencies since loopy belief propagation is used for inference.

Therefore, we propose a feature-wise hierarchical neural CRF tagger \cite{lample2016neural,ma2016end, yang2016multi} with independent predictions for each coarse-grained feature for a given time-step, without explicitly  modeling the pairwise dependencies.  

\subsubsection{Hierarchical Neural CRF model}
The hierarchical neural CRF model comprises of two major components, an \emph{encoder} which combines character and word-level features into a continuous representation and a multi-class multi-label \emph{decoder}. Given an unlabeled sequence $x$, the \emph{encoder} computes the context-senstive hidden representations for each token $x_i$. These representations are shared across  $|F|$ independent linear-chain CRFs for inference. We refer to this model as \textsc{mdcrf}. 

\paragraph{Decoder:} Our decoder comprises of  $|F|$ independent feature-wise CRFs whose objective function is given as follows:
\[ p(\mathbf{y}|\mathbf{x}) =  \prod_{j=1}^{F} p_f(\mathbf{y_f} | \mathbf{x} ) \]
\[ p_f(\mathbf{y_f} | \mathbf{x} ) = \frac{\prod_{t=1}^{n} \psi_i(y_{f,t-1},y_{f,t},\mathbf{x},t)}{Z(\mathbf{x})} \]
where $F$ = \{POS, Case, Gender,...\} is the set of coarse-grained features observed in the training dataset.  $p_f(\mathbf{y_f} | \mathbf{x})$ is a feature-wise  CRF tagger  with $\psi_i(y_{t-1},y_t, \mathbf{x})=\exp(\mathbf{W_f}^{T}_{y_{f,t-1},y_{f,t}}\mathbf{x}_i + \mathbf{b_f}_{y_{f,t-1},y_{f,t}})$ being the energy function for each feature $f$. During inference the predictions from each feature-wise decoder is concatenated together to output the complete morphological analysis of the sequence $x$.

\paragraph{Encoder:}

We adopt a standard hierarchical sequence encoder which is shared among all the $|F|$ feature-wise decoders. It consists of a character-level bi-LSTM that computes hidden representations for each token in the sequence. These subword representations help in capturing information about morphological inflections. To further enforce this signal, we add a layer of self-attention \cite{vaswani2017attention} on top of the character-level bi-LSTM. Self-attention provides each character with a context from all the characters in the token. A bi-LSTM modeling layer is added on top of the self-attention layer which produces a token-level representation. These representations are then concatenated with a word embedding vector and fed to another bi-LSTM to produce context sensitive token representations which are then fed to all the $|F|$ CRFs for inference.

\subsubsection{Adding Linguistic Knowledge} 
Part-of-speech (POS) is perhaps the most important coarse-grained feature. Not only is every token annotated for POS, but most other features depend on it. For instance, verbs do not have Case, nouns do not have Tense. 
In order to leverage these linguistic constraints, we incorporate POS information for each token into our shared encoder. We refer to this variant of the model as \textsc{mdcrf+pos}, as shown in Figure \ref{fig:model}.

Since POS tags are not available as input, we first run a separate  hierarchical neural CRF tagger for POS alone and use the model predictions as input to the \textsc{mdcrf+pos}.   For each token, we encode its predicted POS tag into a continuous representation and concatenate it with the character and word-level token representations. Finally, these concatenated representations are fed to the word-level bi-LSTM and inference is performed using $|F|$-1 decoders, excluding the POS decoder.  Going forward, we use this model architecture for all our experiments unless otherwise noted.  

\begin{figure}
 \centering
  \includegraphics[width=0.5\textwidth]{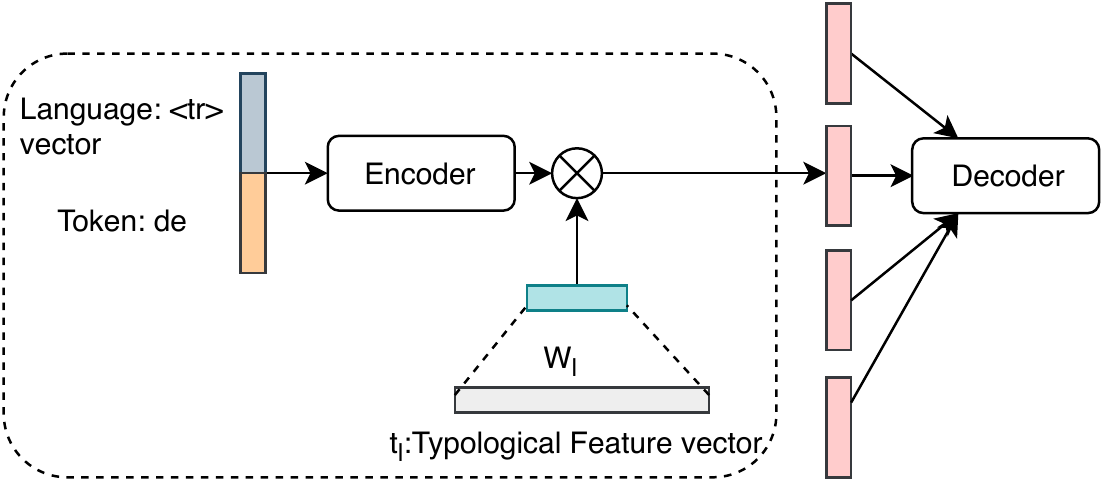}
  \caption{Polyglot model being used for the token ``de" in Turkish, denoted by language vector $<$tr$>$.}
  \label{fig:polyglotmodel}
\end{figure}

\subsubsection{Multi-lingual Transfer}
So far, we have described our model architecture for a monolingual setting. However, the performance of neural models is highly dependent on the availability of large amounts of annotated data, making it challenging to generalize to low-resource languages. Cross-lingual transfer learning attempts to alleviate this challenge by transferring knowledge from high-resource languages. Prior work \cite{cotterell2017cross, malaviya2018neural, buys2016cross} has shown the benefits of cross-lingual transfer for morphological tagging. \citet{malaviya2018neural} restrict to transferring from one language, whereas \citet{cotterell2017cross} show that multi-source transfer performs better than single-source. Inspired by this, we experiment with two approaches for multi-lingual transfer learning.

\paragraph{\textsc{multi-source}:} In this method, we augment the training data from related languages with the target language data. Similar to \citet{cotterell2017cross}, we perform a hard clustering of languages based on the typological and orthographic similarity of the source languages with the target language. For instance, we construct a language cluster Indo-Aryan, which comprises of all the languages in the dataset that belong to the Indo-Aryan language family which are Hindi, Marathi and Sanskrit. For some larger language families such as Germanic and Slavic,  we construct language clusters from a subset of languages. For instance, the North-Germanic language cluster comprises of treebanks from German, Norwegian, Swedish and Danish. Some languages such as Urdu, Tamil are the only representative languages of their respective language families in the dataset. For these languages, we create a cluster with the next closest language with respect to typology or orthography.  For Urdu, we add Hindi because of typological similarity. For other such isolates, we add Turkish because of its extensive agglutination. A total of 24 language clusters were defined based on the literature and with help from a linguist, the details of which can be seen in the Appendix Section \S \ref{lang}.

Given a language cluster,  all the training data from each language within it is first concatenated together. Then, for each language  we concatenate the language embedding vector with the token representation in the encoder by adding the language id \textsc{\small{$<$lang id$>$}} at the beginning and end of each sequence.  Given a sequence $\mathbf{x}$, the encoder produces contextualized hidden representation $h_i$ for each token $x_i$:
\[ h_i = W_{encoder}(e_i, c_i, p_i, l_i) \] where $e_i$ is the word embedding vector, $c_i$ is the character-level representation, $p_i$ is the POS embedding and $l_i$ is the language embedding vector. This is done to help the model disambiguate languages as often same tokens have different morpho-syntactic description across languages. For example, the token ``\token{}" is a part of both Hindi and Marathi vocabulary. In Hindi it denotes a CONJ whereas in Marathi it is a pronoun with the following description: 3;MASC;PRO;NOM;SG.

\paragraph{\textsc{polyglot}:} Languages are often related to multiple languages along different dimensions. For instance, Swedish is lexically similar to German, but it is morpho-syntactically closer to English. 
% We hope that feeding explicit typological information to the encoder will assist the model in capturing this intuition, for which we take inspiration from the polyglot model proposed by \citet{tsvetkov2016polyglot}. 
To enable a model to utilize these relationships, we feed explicit typological information to the encoder, drawing inspiration from the polyglot model proposed by \citet{tsvetkov2016polyglot}. 
In this multilingual model, we first concatenate all the training data from the source languages, similar to the \textsc{multi-source} setting and compute  $h_i$ for each token. 
Then context vector $h_i$ is factored by the typology feature vector $t_l$ to integrate these manually defined features as follows:
\[ f_l = \tanh(W_lt_l+ b_l) \]
\[g_{i}^{l} = h_i \otimes f_l^T\]
where $W_l, b_l$ are language-specific parameters which project the typology vector into a low-dimentional space. Finally, $g_{i}^{l}$ computes the global-context language matrix which is vectorized into a column vector and fed to the decoder, as shown in Figure \ref{fig:polyglotmodel}.

\citet{tsvetkov2016polyglot} derive their typology vectors from the URIEL database \cite{littell2017uriel}. We consider a subset of these typology features which are most relevant to the task of morpho-syntactic analysis and obtain 18 Syntax-WALS features.\footnote{\small{S-SVO, S-SOV, S-VSO, S-VOS, S-OVS, S-OSV, S-SUBJECT-BEFORE-VERB, S-SUBJECT-AFTER-VERB, S-OBJECT-AFTER-VERB, S-OBJECT-BEFORE-VERB, S-SUBJECT-BEFORE-OBJECT, S-SUBJECT-AFTER-OBJECT,S-ADPOSITION-BEFORE-NOUN, S-ADPOSITION-AFTER-NOUN, S-POSSESSOR-BEFORE-NOUN, S-POSSESSOR-AFTER-NOUN, S-ADJECTIVE-BEFORE-NOUN, S-ADJECTIVE-AFTER-NOUN}} However, we observed that for most language clusters, these typology feature values within a cluster were not discriminating, which defeats the purpose of using \textsc{polyglot} for disambiguating languages across typological dimensions. Therefore, we construct custom typological vector per each language cluster based on the training data global statistics. 

For every coarse-grained feature, this constructed vector contains the proportion of words in the training data that are annotated with that feature. We also experiment with calculating these proportions separately for words for each POS label (N, V, ...). Given the importance of POS, we also include the number of fine-grained POS labels that the most frequent coarse-grained features ({Gender, Number, Person, Case}) can take. This results in bi-gram features such as N-FEM, N-NOM, N-SG. 
% These features are then condensed to exclude those that do not occur in the entire language cluster, to avoid sparse features.
We remove features which do not occur within a given cluster to avoid sparse features. Table \ref{tab:manual} shows a portion of the example vector constructed for the Indo-Aryan cluster. From the table we can see that, some features such as ADJ-Gender-FEM and V-Person-1 are present in all the three languages within the cluster. Whereas some features such as ADJ-Gender-NEUT  is absent from Hindi because Hindi only has two genders which are MASC and FEM.

\begin{table}[h]
\small
 \begin{center}\resizebox{0.5\textwidth}{!}{
 \begin{tabular}{c|c|c|c}
 \textbf{Feature} & \textbf{Hindi} & \textbf{Marathi} & \textbf{Sanskrit}\\
 \toprule
 ADJ-Gender-FEM & 0.054 & 0.144 & 0.080 \\
 V-Person-1 & 0.004 & 0.037 & 0.0736\\
  ADJ-Gender-NEUT   & 0.0 & 0.144 & 0.159 \\
  ADJ-Case-{DAT/GEN} & 0.0002 & 0.0 & 0.0 \\
 \bottomrule
 \end{tabular}
 }
 \caption{Example of manually constructed typology features for the Indo-Aryan cluster.}
%   \vspace{-1em}
   \label{tab:manual}
 \end{center}
 \end{table}

\paragraph{Training Regime:} For both the multi-lingual transfer methods,  we train one model per language cluster and fine-tune this model for each individual language. which saves time and compute for training 107 individual models from scratch. Furthermore, since a language cluster can have multiple high-resource languages, we take \textit{min (5000, \#training data-points)} for each language to have a tractable training time. We up-sample the low-resource languages to match the number of training data-points of the high-resource languages. %These clusters were created based on the typological similarity between languages, details of which are in the Appendix.

\section{Contextual Lemmatization}

We use the neural model from \citet{malaviya2019simple} for contextual lemmatization. 
This is a neural sequence-to-sequence model with hard attention, which takes both the inflected form and morphological tag set for a token as input and produces a lemma, both at the character level.
The decoder uses the concatenation of the previous character and the tag set to produce the next character in the lemma. 
The lemmatization model is jointly trained with an LSTM-based tagger using jackknifing to reduce exposure bias in training: \citet{malaviya2019simple} report significantly lower lemmatization results training with gold tags and using predicted tags only at test time. 
We use their tagger for training and our contextual morphological analysis models' predicted tags at evaluation time. 
This model served as the baseline lemmatizer for Task 2; we refer readers to the shared task paper for model details \cite{sigmorphon2019}. 

%% file: experiments.tex
\begin{table*}
\small
 \begin{center}\resizebox{\textwidth}{!}{
 \begin{tabular}{c|c|c|c|c|c|c|c}
 \multirow{2}{*}{\textbf{Language}} & \multirow{2}{*}{\textbf{Model}} & \multicolumn{3}{c|}{\textbf{tgt-size=100}}&\multicolumn{3}{c}{\textbf{tgt-size=1,000}} \\
 \cmidrule(lr){3-5} \cmidrule(lr){6-8} 
 & & Accuracy & F1-Macro & F1-Micro & Accuracy & F1-Macro & F1-Micro
\\
 \toprule
 RU/BG & \textsc{\small{mdcrf + pos + multi-source}}  & \textbf{69.13} & \textbf{85.78} & \textbf{85.86}  & \textbf{82.72} &	\textbf{92.15} & \textbf{92.17} \\
 & \cite{malaviya2018neural} & 46.89 & 64.75 & 64.46 & 67.56 &	82.06 &	82.11 \\
  & \cite{cotterell2017cross} & 52.76 & 58.23 & 58.41 & 71.90 &	77.89 &	77.97\\
 \midrule
 FI/HU & \textsc{\small{mdcrf + pos + multi-source}} & \textbf{57.32}  & \textbf{80.11} & \textbf{78.86}  & \textbf{70.24} &	\textbf{85.44} &	\textbf{84.86} \\
  & \cite{malaviya2018neural} & 45.41 & 68.63 & 68.07 & 63.93 &	85.06 &	84.12 \\
  & \cite{cotterell2017cross} & 51.74 & 68.15 & 66.82 &61.8 &	75.96 &	76.16  \\
 \midrule
 \end{tabular}
 }
 \caption{Comparing our model for bilingual transfer with previous baselines.}
   \label{tab:baseline1}
 \end{center}
 \end{table*}
\section{Experiments}
\label{exp}
% To examine the effectiveness of our approach for contextual morphological analysis, 
We conduct the following experiments:  
We compare our multi-lingual transfer approach with the baselines \citet{malaviya2018neural} and \citet{cotterell2017cross} under the same experimental settings. 
Next, we compare our approach with the shared task baseline \cite{sigmorphon2019}. 
Finally, we analyze the contributions of different components of our proposed method. 
% We now describe the data pre-processing steps we followed for all the experiments.

\paragraph{Baselines:}
\citet{cotterell2017cross} formulate this task as a sequence prediction problem with the output space being the set of all possible tagsets seen in the training data. Specifically, they construct a neural network based multi-class classifier where each tagset \{\textit{N;PL;NOM;FEM}\} forms a class. %The objective function is shown below:
%\[ p(\mathbf{y}|\mathbf{x}) = \prod_{i=1}^{n} p(y_i | \mathbf{x}) \]
Since the output space is only restricted to the tagsets seen in the training data, this method cannot generalize to unseen tagsets. Furthermore, for morphologically rich languages such as Russian or Turkish, the output space of the tagset is huge leading to sparse training data. \cite{sigmorphon2019} follow a similar approach.

To overcome these drawbacks \citet{malaviya2018neural} consider a feature-wise model which predicts fine-grained labels for corresponding coarse categories \{POS,Case,...\}. Since morpho-syntactic properties are often correlated, they model these inter-dependencies using a factorial CRF and define two inter-dependencies: 1) a \emph{pairwise} dependency, which models correlations between the morpho-syntactic properties within a token, and 2) a \emph{transition} dependency, which models label correlations across all tokens in a sequence. %The objective function is shown below:
%\[ p(\mathbf{y}|\mathbf{x}) = \frac{1}{Z(\mathbf{x})} \prod_{i=1}^{n} \prod_{\alpha \in C} \psi_{\alpha} (\mathbf{y_{\alpha}}, \mathbf{x}, i) \]
%where C is the set of factors comprising of both pairwise and transition dependencies, $\mathbf{y_{\alpha}}$ is the assignment to the subset of variables whose dependency is being modeled.  
Although this formulation provides the flexibility to produce any combination of tagsets, this model is computationally expensive to train since the factors model dependencies between all labels of all coarse-grained features, leading to \textgreater 20k factors.
\paragraph{Data processing:}
We use the train/dev/test split provided in the shared task \cite{mccarthy-etal-2018-marrying}.\footnote{\url{https://github.com/sigmorphon/2019/tree/master/task2}} 
Since we model feature-wise prediction for each coarse-grained feature, our model requires the provided data to be annotated for coarse-grained features. Therefore, we construct a feature-label dictionary based on the UM documentation\footnote{\url{https://unimorph.github.io/doc/unimorph-schema.pdf}} to map the individual fine-grained traits, which are in the UM schema, to their respective coarse-grained categories. This transforms the  tagset \{N;PL;NOM;FEM\} as \{\textit{POS=N;Number=PL;Case=NOM;Gender=FEM}\}. We note that usually a token has a subset of the coarse-grained categories, therefore  we extend the morphological tagset for each token by adding the remaining features observed in the training set  and assigning them a special value ``\textit{\_}'' which denotes null.  

\paragraph{Hyper-parameters:} We use a hidden size of 200 for each direction of the LSTM with a dropout of 0.5. For the character-level bi-LSTM we use a hidden size of 25. We use 100 dimentional size for word and language embeddings with  64 dimensional POS embeddings, all randomly initialized. SGD was used as the optimizer with learning rate of 0.015. The models were trained until convergence. For \textsc{Polyglot}, we project the constructed typology vector into 20 dimension hidden size.

%% file: Results.tex
\section{Results and Discussion}
\label{results}

Table~\ref{tab:baseline1} shows the comparison results of our proposed approach with the baselines \cite{malaviya2018neural,cotterell2017cross} using cross-lingual transfer. Here \textsc{mdcrf+pos} refers to our model architecture and \textsc{multi-source} refers to our   multi-lingual transfer approach. \citet{malaviya2018neural} and \citet{cotterell2017cross} test their approach on UD v2.1 \cite{nivre2017universal} under two settings: \emph{tgt size} $=100$ and \emph{tgt size} $=1000$, where \emph{tgt size} denotes the number of target language data-points used during training. \citet{malaviya2018neural} transfer from one related high-resource language. We use the same experimental resources for comparison and for a fair comparison we do not fine-tune on the target language. Of the four language pairs tested by \citet{malaviya2018neural}, we choose RU/BG and FI/HU for comparison, where BG  and HU are the target languages and RU  and FI  are the respective transfer languages, since these languages are morphologically challenging.  We see that under both settings our approach outperforms the baselines by a significant margin for both the language pairs.

Next, we compare our multi-lingual transfer approaches \textsc{multi-source} and \textsc{multi-source + polyglot} in order to decide the model for our final submission. We conduct experiments on three low-resource languages: Marathi (\emph{mr-ufal}), Sanskrit (\emph{sa-ufal}) and Belarusian (\emph{be-hse}), all of which have $<400$ training data-points. The italicized text denotes the treebank used in the experiments. For \emph{mr-ufal} and \emph{sa-ufal}, we transfer from a related high-resource language of Hindi (\emph{hi-hdtb}). For \emph{be-hse}, we transfer from two related languages, Russian (\emph{ru-gsd}) and Ukrainian (\emph{uk-iu}). However, from Table \ref{tab:polyglot}, we see that the performance of the two models is comparable. Therefore, for our final submission we use only \textsc{multi-source} which is much faster to train than the  \textsc{multi-source + polyglot}. 
% We discuss the reasons for this comparative performance in detail in Section \S{\ref{analysis}}.
We discuss their comparative performance in greater detail in Section \S{\ref{analysis}}.
\begin{table}[h]
\small
 \begin{center}\resizebox{0.5\textwidth}{!}{
 \begin{tabular}{c|c|c|c}
 \textbf{Model} & \textbf{mr-ufal} & \textbf{sa-ufal} & \textbf{be-hse}\\
 \toprule
 \textsc{\small{multi-source}} & \textbf{63.52 / 78.22}   & 42.78 / \textbf{67.64} & \textbf{77.07} / 82.89 \\
  \textsc{\small{+polyglot}} & 61.18 / 77.42  & \textbf{43.81} / 65.94  & 76.51 / \textbf{83.27} \\
 \bottomrule
 \end{tabular}
 }
 \caption{Multi-lingual comparison results for Marathi (\emph{mr-ufal}), Sanskrit (\emph{sa-ufal}) and Belarusian (\emph{be-hse}) on the validation set.}
%   \vspace{-1em}
   \label{tab:polyglot}
 \end{center}
 \end{table}

Finally, we compare our approach with the shared task baseline. Table \ref{tab:result1}, \ref{tab:result2} in the Appendix shows our results for all 107 treebanks. We observe that out system achieves an average improvement of +14.70 (accuracy) and	+4.63 (F1)  over the provided baseline \cite{sigmorphon2019}. We note that for the shared task submission, we did not use self-attention over the character-level representations. Therefore, we additionally show the results after adding self-attention. We observe that the addition gives an average improvement of 
+0.60 (accuracy) and +0.30 (F1) over our previous best submission.

\subsection{Analysis}
\label{analysis}
Here we analyze the different components of our model in an effort to understand what it is learning.

\paragraph{Why does adding POS help?} 
As discussed earlier (\S \ref{method}), we explicitly add the POS feature in the form of embeddings into the shared encoder. To evaluate the contribution of POS alone, we conduct monolingual experiments without concatenating the POS embeddings with the token-level representations. Table \ref{tab:pos} outlines the ablation results for three treebanks with varying training size. We observe that our monolingual model \textsc{mdcrf} significantly outperforms the baseline \cite{sigmorphon2019} by +13.72 accuracy and +3.82 F1 (avg). On adding POS, we further gain +3.56 accuracy and +0.71 F1 over \textsc{mdcrf} across the three treebanks. We note that this improvement is more pronounced for the low-medium resource languages of Marathi (+6.12 accuracy) and Ukrainian (+3.57 accuracy). 
\begin{table}[h]
\small
 \begin{center}\resizebox{0.5\textwidth}{!}{
 \begin{tabular}{c|c|c|c}
 \textbf{Model} & \textbf{mr-ufal} & \textbf{uk-iu} & \textbf{hi-hdtb}\\
 \toprule
 \textsc{\small{MDCRF+POS}} & \textbf{64.71 / 79.40}   & \textbf{84.79 / 92.03} & \textbf{90.46} / 96.69 \\
  \textsc{\small{MDCRF}} & 58.59 / 77.91  & 81.22 / 91.35  & 89.45 / \textbf{96.73} \\
   \citet{sigmorphon2019} & 43.76 / 73.38 & 63.36 / 87.01 & 80.96 / 94.14 \\
 \bottomrule
 \end{tabular}
 }
 \caption{Ablation results for Marathi (\emph{mr-ufal}), Ukrainian (\emph{uk-iu}) and Hindi (\emph{hi-hdtb}) with training size of 373, 5441, 13381 respectively on the validation set.}
   \label{tab:pos}
 \end{center}
 \end{table}
 
\begin{figure}[!h]
\begin{center}\resizebox{0.5\textwidth}{!}{
    \includegraphics{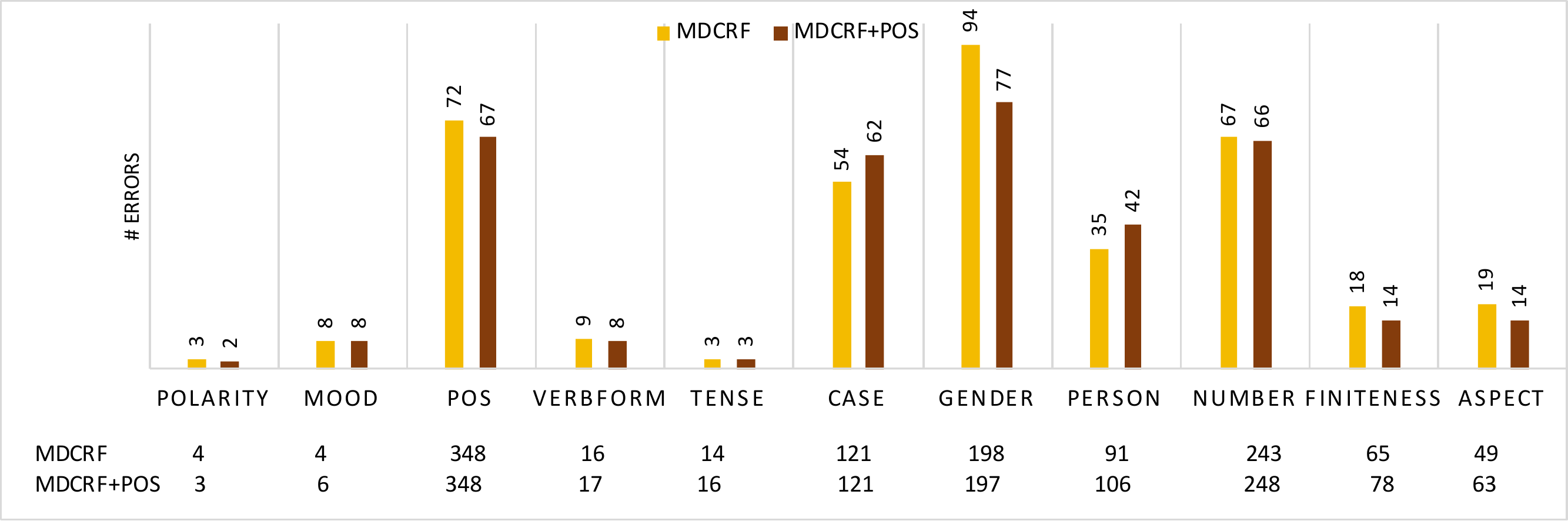}
    }
    \end{center}
    %\vspace{-0.5em}
    \caption{Number of errors per coarse-grained feature for Marathi comparing the addition of POS to the encoder. The rows at the bottom denote the total number of predictions per each feature for both the models.}
    \label{fig:pos}
\end{figure}

To understand where the addition of POS helps, we analyse the number of errors made per each coarse-grained feature. For the example of Marathi, POS helped the most in reducing Gender errors (Figure \ref{fig:pos}). For some word forms, the gender may be inferred from inflectional form alone, but for others, this information may be insufficient, e.g. ``\noun{}" (price.N.FEM.SG.ACC) in Marathi which does not have the traditional female suffix ``\suffix{}". 
We observe that this behavior corresponds to POS: verbs and adjectives are more predictable from surface forms alone than nouns. 
The addition of POS information in the encoder helps the model learn to weigh different encoded information more heavily when assigning gender to different parts of speech. For Ukrainian and Sanskrit, POS information also helped reduce errors in Case and Number. More details can be found in Appendix Section \S \ref{langanalysis}.

\citet{tkachenko2018modeling} also model dependence on POS with a POS-dependent context vector in the decoder. However, they observe no significant improvement; we hypothesize that incorporating POS information into the shared encoder instead provides the model with a stronger signal.
\begin{figure*}%
    \centering
    \subfigure[Marathi]{%
    \label{marathi}%
    \includegraphics[width=0.33\textwidth, trim={4cm 0 4cm 0},clip]{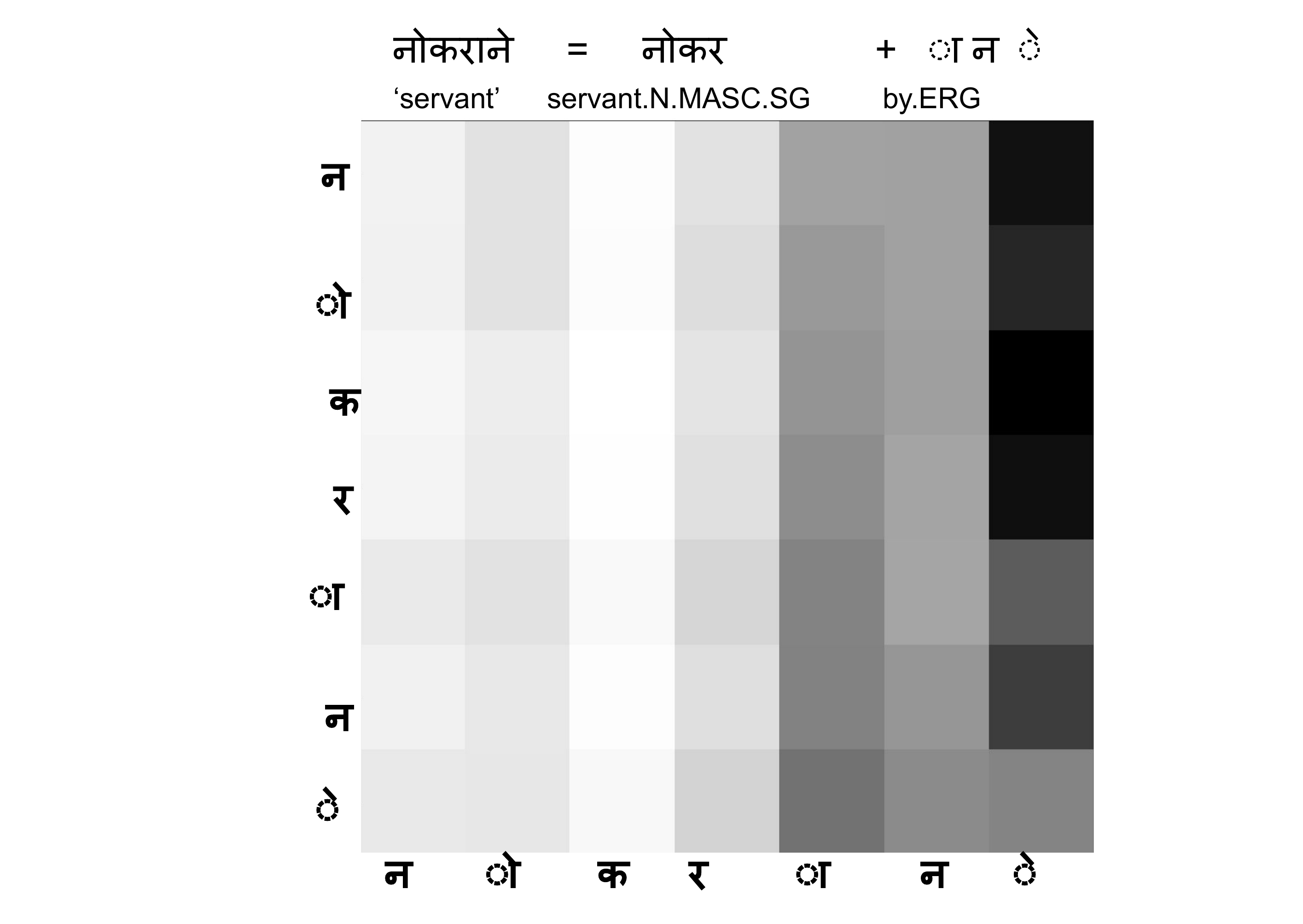}}%\\
    \subfigure[Indonesian]{%
    \label{indonesian}%
     \includegraphics[width=0.33\textwidth, trim={7cm 1cm 5cm 0},clip]{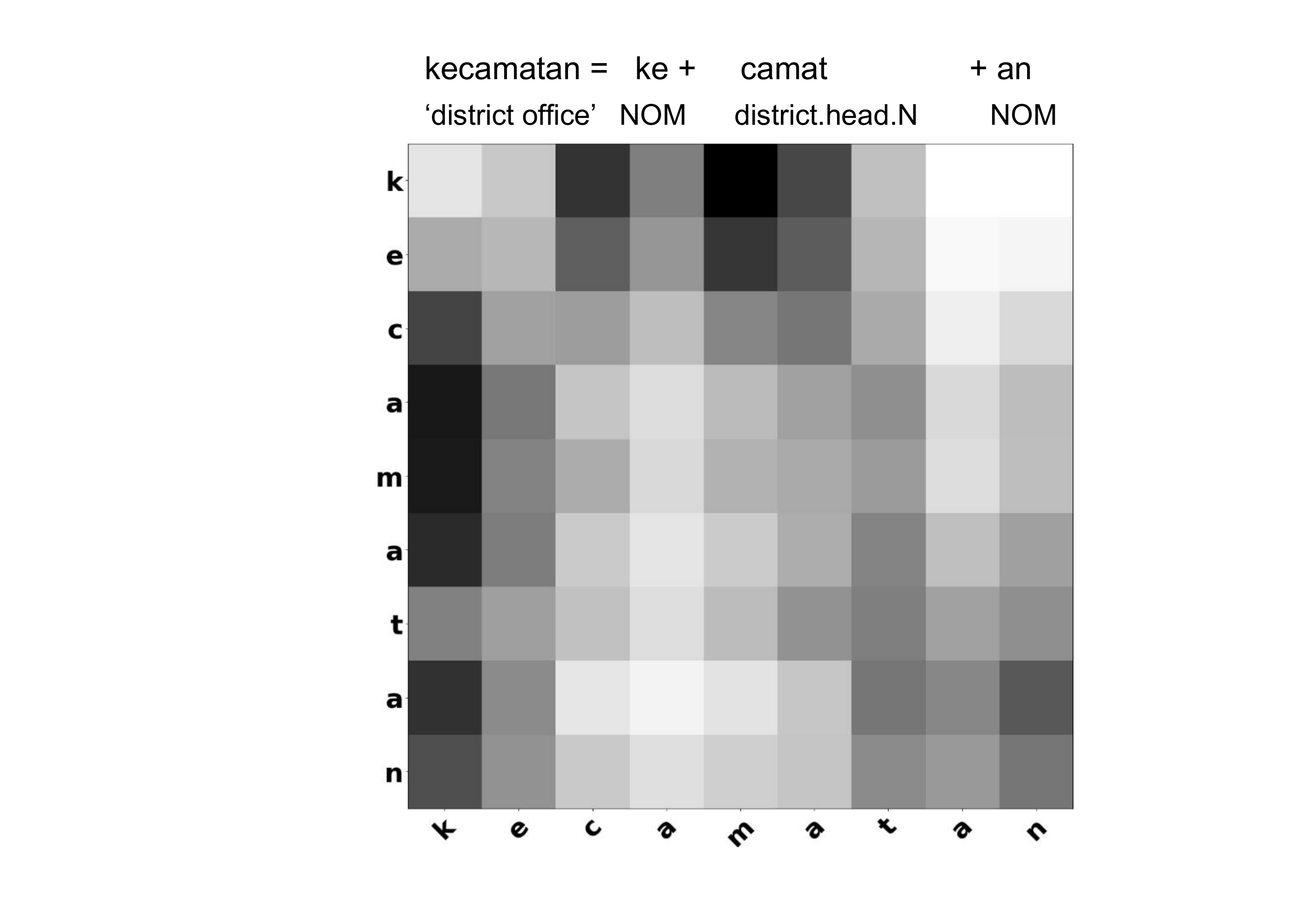}}%\\
    \subfigure[Belarusian]{%
    \label{be}%
    \includegraphics[width=0.33\textwidth, trim={7cm 1cm 3.5cm 0},clip]{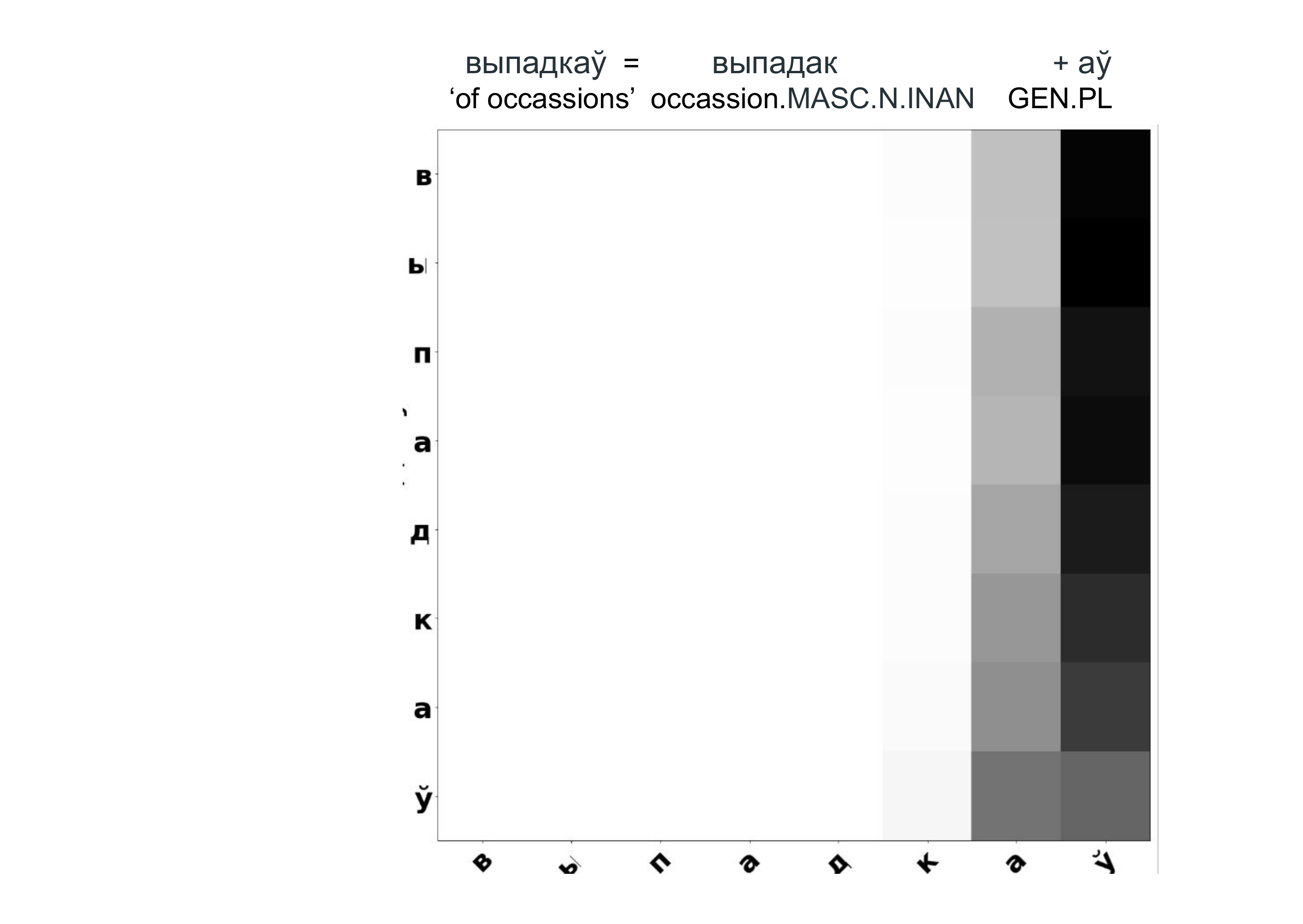}}%\\
~
  \caption{Character-level attention maps for three typologically different languages. Marathi and Belarusian display morphological inflections pre-dominantly as suffix. Indonesian displays inflections in the form of prefix, suffix and circumfix where the affix is found both at the beginning and end of a token.}%
\label{fig:eg4}%
\end{figure*}

\paragraph{What is the model learning?} One of the major advantages of our model's use of self-attention is that it enables us to provide insights into what the model has learned. As seen in Figure \ref{fig:eg4}, we found evidence of the model learning language-specific inflectional properties. Both Marathi and Belarusian display morphological inflections predominantly in the form of suffix and the attention maps for both these languages demonstrate the same. For the Marathi example, the last three characters denote the ergative case and we can see that the attention weights are concentrated on these three characters. Similarly for the Belarusian example, the last two characters denote the genitive case with plural number and is the focus of the attention. For Indonesian, inflections can be also found as circumfixes where the affix is attached at both the beginning and end of the token. For instance, both \emph{ke-} and \emph{-an} affixes are appended to form nouns and we can see from Figure \ref{fig:eg4} that the attention is focused both on the prefix and the suffix. Interestingly for Indonesian, the model seems to have also discovered the stem \emph{camat}, as evidenced from the attention pattern.  

\begin{figure}[h]
\begin{center}\resizebox{0.5\textwidth}{!}{
    \includegraphics[trim={0 1cm 0 0},clip]{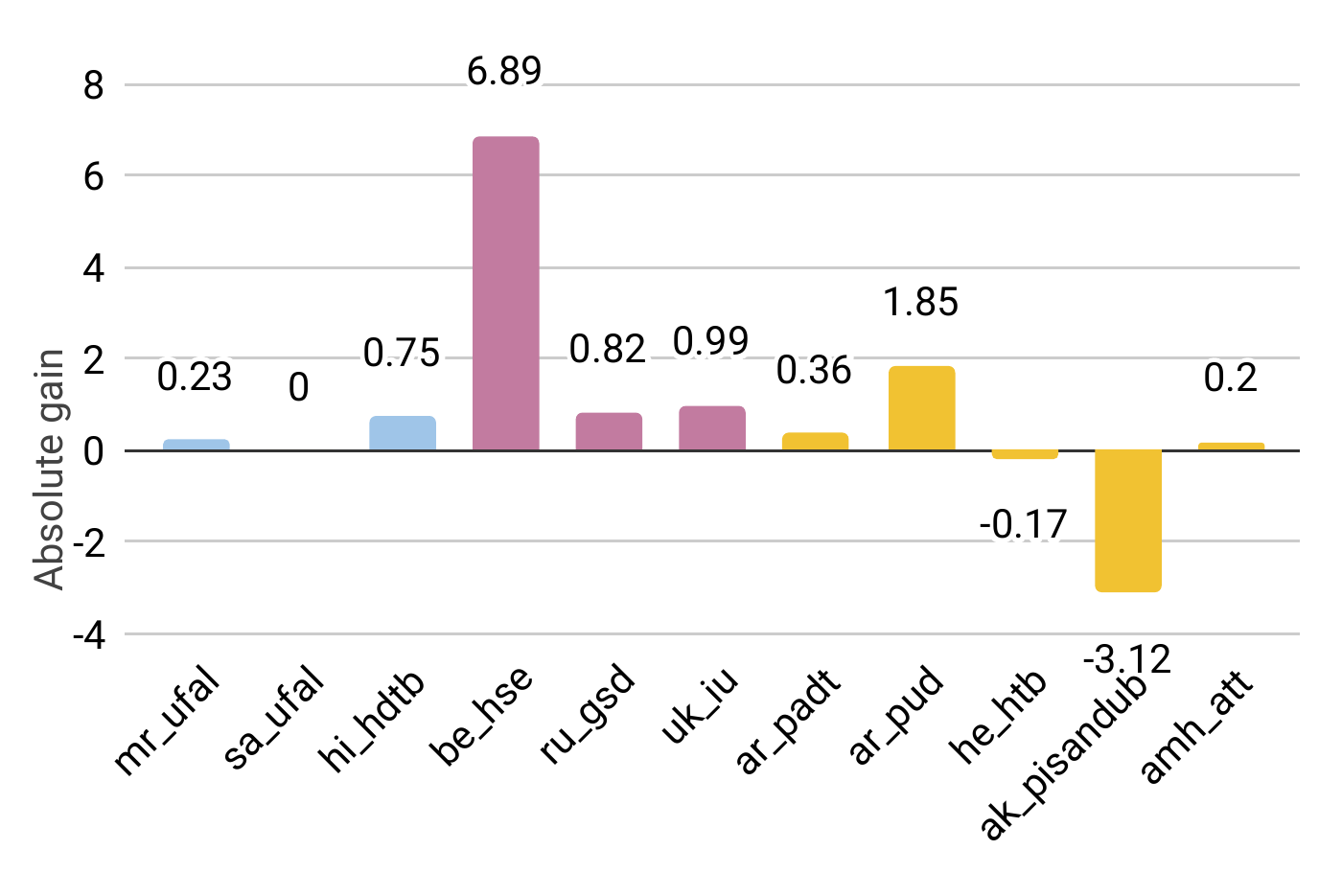}
    }
    \end{center}
    \caption{Absolute gain of multi-lingual transfer over monolingual models. Blue denotes the \textit{Indo-Aryan} cluster, pink the  \textit{Slavic}, and yellow the \textit{Semitic}.}
    \label{fig:dist}
\end{figure}

\paragraph{Does \emph{time-depth} matter for transfer learning?} ~\\
As discussed earlier, we train one model per language cluster for multi-lingual transfer learning. 
% We ask the question ``does the \emph{time-depth} of languages affect the extent of transfer?" 
We compare different clusters to see if \emph{time-depth} of the languages within a cluster affects the extent of transfer. \emph{Time depth} is the period of time that has elapsed since all languages in the group were a single language (in other words, the time since divergence).
We consider the following three clusters: Hindi-Marathi-Sanskrit (Indo-Aryan), Russian-Ukrainian-Belarusian (Slavic) and Arabic-Hebrew-Amharic-Akkadian (Semitic). These three clusters were chosen because the languages in them became separate languages at varying time-depths.  For instance, in the Semitic cluster the languages diverged roughly 5000 years ago, whereas for the Slavic cluster the time-depth is $<$1000 years. Therefore, we expect transfer to help more for languages where the time-depth is more recent. In Figure \ref{fig:dist}, we compare the \textsc{{multi-source}} model with our best mono-lingual model \textsc{{mdcrf+pos}} and we see that transfer helps most for the Slavic cluster by +2.9 accuracy. For the Indo-Aryan cluster it helps by +0.32 accuracy and for the Semitic cluster we observe a slight negative effect with transfer (-0.0176 accuracy). This supports our hypothesis that  \emph{time-depth} does affect the extent of transfer learning with language clusters having lower  \emph{time-depths} benefiting the most.  

One particular advantage that the Slavic cluster has over both the Indo-Aryan and Semitic clusters is the similarity of script.  Russian, Belarusian, and Ukrainian use variants of the same script; Hindi, Sanskrit, and Marathi do, as well, but the Semitic languages all use different scripts. This is also attributed to the shallower time-depths of the Slavic and Indo-Aryan clusters. Therefore, as suggested by the anonymous reviewers, we add Czech and Polish to the Slavic cluster and see to what extent the scripts are confusing the model. Czech and Polish use different script as compared to Russian, Belarusian, and Ukrainian.    We observe that \textsc{{multi-source}} model like before, achieves  similar improvements over the monolingual models for Belarusian (+8.17 accuracy) and Ukrainian (+1.2 accuracy). However, a slight decrease is observed for Russian ( -0.45 accuracy). This suggests that the \textsc{{multi-source}} model is robust to scriptal changes and benefits the low-resource languages  by learning from typologically similar languages, more so for language clusters with shallow time-depths.
 \begin{figure*}%
    \centering
    \subfigure{%
    \label{case_dataaug}%
    \includegraphics[width=0.5\textwidth, trim={7cm 0 0 3cm},clip]{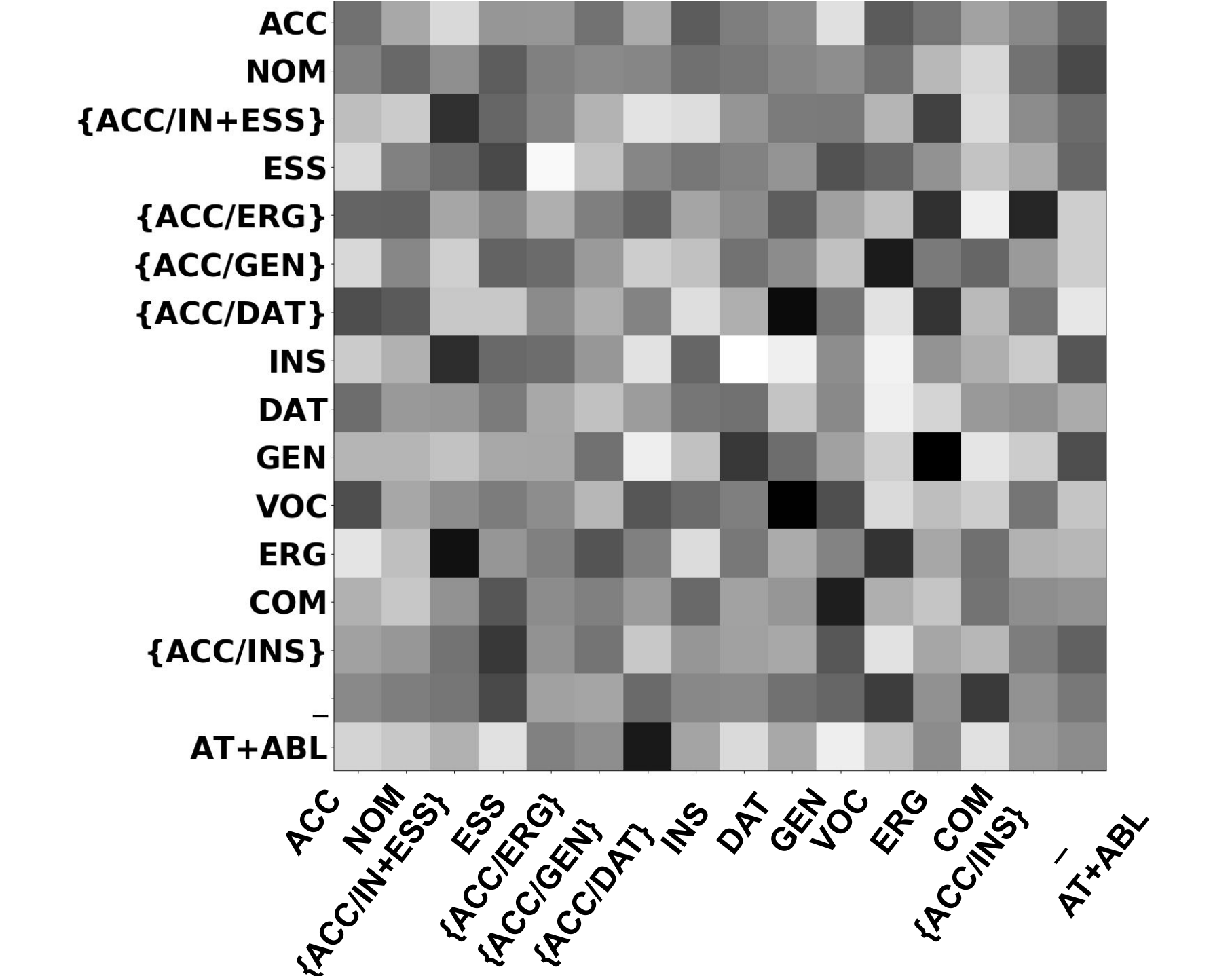}}%\\
    \subfigure{%
    \label{case_polyglot}%
     \includegraphics[width=0.5\textwidth, trim={7cm 0 0 2.5cm},clip]{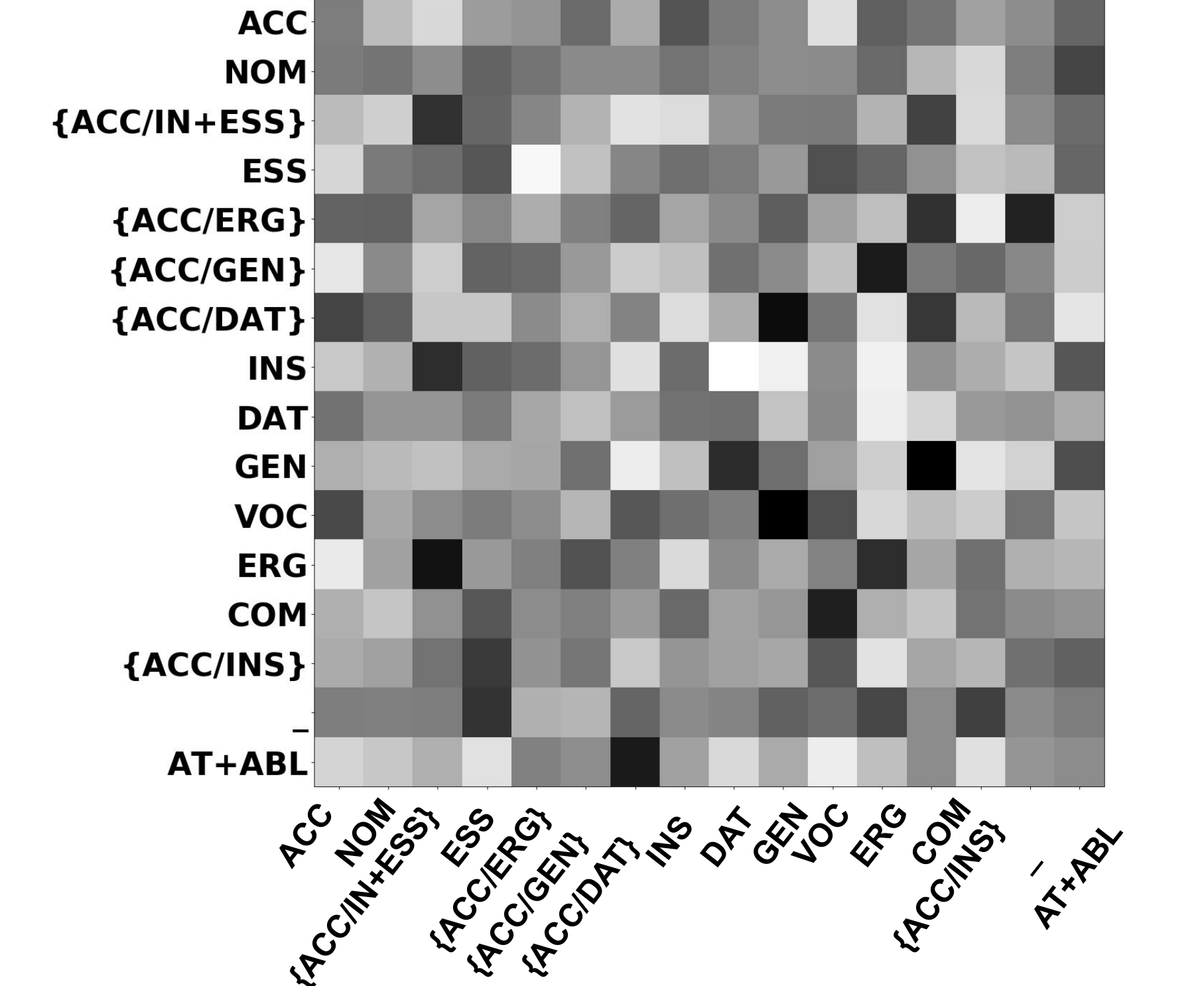}}%\\
~
  \caption{Transition weights for the \emph{Case} feature for Hindi across  \textsc{\small{multi-source}} (left) and \textsc{\small{multi-source + polyglot}} (right) models trained with Hindi (\emph{hi-hdtb}), Marathi  (\emph{mr-ufal}) and Sanskrit  (\emph{sa-ufal}).}%
\label{fig:eg5}%
\end{figure*}

\paragraph{Why did \textsc{{polyglot}} not help further?} We hypothesize that one reason why \textsc{{polyglot}} did not help over \textsc{{multi-source}} is because the language embedding vector probably learns the same typological information which the typology vector encodes. Hence, the typological vector doesn't seem to add any new information.  As evidence, we look at the transition weights learned in both the models; as shown in Figure \ref{fig:eg5}, we see that the transition weights learned for the Case feature are very similar for both \textsc{{multi-source}} and \textsc{{multi-source + polyglot}}. In the future, we plan to explore the contextual parameter generation method \cite{platanios-etal-2018-contextual} for leveraging the typology vectors to inform the decoders during inference.

\subsection{Error Analysis}
In this section, we analyze the major error categories for the \textsc{{multi-source}} model for the Indo-Aryan cluster. We observe that Gender, Case, Number, Person features account for the most number of errors (65\% for Marathi, 49\% for Sanskrit). One reason for this is the non-overlapping output label space across the languages within a cluster. For instance, in the Indo-Aryan cluster, Hindi is a high-resource language ($> 13k$ training sentences) with Marathi (373) and Sanskrit (184) being the low-resource languages. We observe that the label space for Case, Gender, Number overlap the least among the three languages. Marathi and Sanskrit have three genders: \emph{NEUT, FEM, MASC} whereas Hindi only has \emph{FEM, MASC}. Furthermore, only two Hindi Case labels (\emph{ACC, NOM}) overlap with Marathi and Sanskrit because in Hindi the labels often have alternatives such as \emph{ACC/ERG, ACC/DAT}. These differences in the output space negatively affect the transfer. For the Slavic cluster, we observe that almost all the feature labels overlap nicely for the languages therein, which is probably another reason why we see a gain of +6.89 for Belarusian in Figure \ref{fig:dist} and only +0.32 increase for Marathi. 

We also note that for some languages such as Belarusian and Russian, the POS errors increased by 25.3\% and 4.4\% respectively for the \textsc{mdcrf+pos} model. This suggests that decoupling POS feature from the other feature decoders harmed the model. In future, we plan to improve the \textsc{mdcrf+pos} model by jointly training POS decoder with the other feature decoders which use the latent representation of POS in an end-to-end fashion.

%% file: conclusion.tex
\section{Conclusion and Future Work}
\label{conclusion}
%We presented our system for contextual morphological analysis. 
We implement a hierarchical neural model with independent decoders for each coarse-grained morphological feature and show that incorporating POS information in the shared encoder helps improve prediction for other features. Furthermore, our multi-lingual transfer methods not only help improve results for related languages but also eliminate the need of training individual models for each dataset from scratch. In future, we plan to explore the use of pre-trained  multi-lingual word embeddings such as BERT \cite{devlin-etal-2019-bert}, in our encoder.

\section*{Acknowledgement}
We are thankful to the anonymous reviewers for their valuable suggestions. 
This material is based upon work supported by the National Science Foundation under Grant No.~IIS1812327.

%% file: AResults.tex
\section{Comprehensive Results}
Table \ref{tab:result1} and \ref{tab:result2} document the comprehensive results of our submissions. \textsc{multi-source} was our previous submission to the shared task. We conducted additional experimentas with the addition of self-attention and also report the results for \textsc{multi-source+self-attention}. We report both the accuracy and F1 metric.

\begin{table*}
\small
 \begin{center}\resizebox{\textwidth}{!}{
\begin{tabular}{c|l|c|c|c|c}
 \textbf{Language} & \textbf{Target} & \textbf{\textsc{\small{multi-source}}} & \textbf{\textsc{\small{multi-source}}}  &\textbf{\cite{sigmorphon2019}} & \textbf{\# Training} \\
 \textbf{Cluster}& & \textbf{\textsc{\small{+ Self-Attention}}} & & &  \textbf{Sentences} \\
  & & \textbf{Accuracy / F1} & \textbf{Accuracy / F1}& \textbf{Accuracy / F1} & \\
 \toprule
armenian & UD-Armenian-ArmTDP & 83.74 / \textbf{88.54} & \textbf{83.83} / 88.17 & - / - & 825 \\
\midrule
austronesian & UD-Indonesian-GSD & \textbf{90.05} / \textbf{93.13} & 90.01 / 93.11 & 71.49 / 86.02 & 4475 \\
\midrule
baltic & UD-Latvian-LVTB & 89.0 / 93.04 & \textbf{89.0} / \textbf{93.08} & 70.21 / 89.53 & 7937 \\
 & UD-Lithuanian-HSE & \textbf{70.29} / \textbf{76.38} & 68.08 / 74.56 & 43.13 / 67.41 & 211 \\
 \midrule
celtic & UD-Breton-KEB & \textbf{85.97} / \textbf{88.78} & 85.07 / 88.07 & 77.41 / 88.58 & 711 \\
 & UD-Irish-IDT & \textbf{76.75} / \textbf{84.1} & 76.5 / 84.11 & 67.45 / 81.72 & 817 \\
  \midrule
dravidian & UD-Tamil-TTB & \textbf{82.92} / 89.91 & 82.48 / 89.77 & 75.64 / \textbf{90.23} & 481 \\
 \midrule
egyptian & UD-Coptic-Scriptorium & 92.02 / 95.28 & \textbf{92.17} / \textbf{95.33} & 87.99 / 93.78 & 673 \\
 \midrule
germanic & UD-Afrikaans-AfriBooms & 96.92 / \textbf{97.37} & \textbf{96.94} / 97.35 & 84.05 / 92.32 & 1548 \\
 & UD-Dutch-Alpino & \textbf{94.85} / \textbf{95.69} & 94.35 / 95.4 & 82.15 / 91.26 & 10867 \\
 & UD-Dutch-LassySmall & 93.48 / 94.08 & \textbf{93.53} / \textbf{94.2} & 76.24 / 88.13 & 5873 \\
 & UD-English-EWT & 94.08 / \textbf{95.46} & \textbf{93.9} / 95.4 & 79.19 / 90.46 & 13298 \\
 & UD-English-GUM & 93.44 / 94.38 & \textbf{93.56} / \textbf{94.47} & 79.63 / 90.04 & 3520 \\
 & UD-English-LinES & \textbf{94.37} / \textbf{95.19} & 93.75 / 94.93 & 81.03 / 90.99 & 3652 \\
 & UD-English-ParTUT & \textbf{92.01} / \textbf{92.69} & 91.95 / 92.61 & 79.57 / 89.04 & 1673 \\
 & UD-English-PUD & 89.41 / 91.42 & \textbf{89.8} / \textbf{91.6} & 78.85 / 88.8 & 801 \\
 & UD-Faroese-OFT & \textbf{80.6} / \textbf{89.27} & 77.52 / 87.87 & 67.11 / 87.27 & 967 \\
 & UD-Gothic-PROIEL & \textbf{84.53} / \textbf{92.93} & 83.0 / 92.47 & 83.01 / 91.3 & 4321 \\
 \midrule
 north- & UD-German-GSD & \textbf{83.72} / \textbf{92.73} & 82.82 / 92.5 & - / - & 12473 \\
  germanic & UD-Danish-DDT & \textbf{91.78} / \textbf{93.72} &  91.34 / 93.61 & 77.89 / 90.89 & 4410 \\
 & UD-Norwegian-Nynorsk & \textbf{94.39} / \textbf{96.35} & 94.29 / 96.33 & 71.8 / 88.16 & 14061 \\
 & UD-Norwegian-NynorskLIA & 93.03 / 94.55 & \textbf{93.75} / \textbf{94.89} & - / - & 1117 \\
 & UD-Swedish-LinES & \textbf{89.92} / \textbf{93.61} & 89.62 / 93.59 & 77.97 / 91.02 & 3652 \\
 & UD-Swedish-PUD & \textbf{87.72} / \textbf{90.01} & 87.13 / 89.8 & 77.78 / 89.32 & 801 \\
  \midrule
hellenic & UD-Ancient-Greek-Perseus & \textbf{84.79} / \textbf{92.1} & 84.27 / 91.88 & - / - & 11136 \\
 & UD-Ancient-Greek-PROIEL & \textbf{88.1} / \textbf{95.55} & 86.01 / 94.67 & - / - & 13665 \\
 & UD-Greek-GDT & \textbf{91.15} / \textbf{96.23} & 90.73 / 96.0 & 78.14 / 93.49 & 2017 \\
   \midrule
indo-iranian & UD-Urdu-UDTB & 77.77 / 92.12 & \textbf{78.05} / \textbf{92.16} & 67.99 / 88.42 & 4105 \\
  \midrule
indoaryan & UD-Hindi-HDTB & 90.76 / 96.77 & \textbf{91.05} / \textbf{96.85} & 80.96 / 94.14 & 13318 \\
 & UD-Marathi-UFAL & \textbf{57.99} / \textbf{73.54} & 57.72 / 73.04 & 43.76 / 73.38 & 373 \\
 & UD-Sanskrit-UFAL & 43.72 / 64.9 & \textbf{46.73} / 68.08 & 44.33 / \textbf{68.34} & 185 \\
   \midrule
isolate & UD-Basque-BDT & \textbf{75.2} / \textbf{88.07} & 75.14 / 87.91 & 67.61 / 87.63 & 7195 \\
  \midrule
italic & UD-Latin-ITTB & \textbf{94.57} / \textbf{97.26} & 94.25 / 97.11 & 77.62 / 93.19 & 16809 \\
 & UD-Latin-Perseus & \textbf{76.17} / \textbf{86.32} & 75.76 / 85.92 & 53.23 / 77.5 & 1819 \\
 & UD-Latin-PROIEL & \textbf{86.78} / \textbf{94.39} & 86.18 / 94.19 & 82.27 / 91.38 & 14721 \\
   \midrule
jako & UD-Japanese-GSD & \textbf{96.8} / \textbf{96.4} & 96.8 / 96.4 & 85.25 / 90.31 & 6557 \\
 & UD-Japanese-Modern & \textbf{95.27} / \textbf{95.32} & 95.27 / 95.32 & 94.29 / 95.2 & 658 \\
 & UD-Japanese-PUD & \textbf{95.94} / \textbf{95.44} & 95.94 / 95.44 & 84.73 / 89.63 & 801 \\
 & UD-Komi-Zyrian-IKDP & \textbf{51.56} / 61.03 & 51.56 / 62.27 & 33.73 / \textbf{62.59} & 70 \\
 & UD-Komi-Zyrian-Lattice & 53.85 / 64.85 & \textbf{54.4} / 65.23 & 45.6 / \textbf{70.61} & 153 \\
 & UD-Korean-GSD & \textbf{92.56} / \textbf{91.68} & 92.56 / 91.68 & 80.18 / 86.08 & 5072 \\
 & UD-Korean-Kaist & \textbf{95.54} / \textbf{94.99} & 95.54 / 94.99 & 84.32 / 89.4 & 21891 \\
 & UD-Korean-PUD & 84.27 / 89.02 & \textbf{84.46} / 89.28 & 81.6 / \textbf{91.15} & 801 \\
 & UD-Kurmanji-MG & 80.82 / 87.79 & \textbf{80.82} / \textbf{87.81} & 70.2 / 85.85 & 604 \\
   \midrule
niger-congo & UD-Bambara-CRB & 91.65 / 94.76 & \textbf{92.41} / \textbf{94.86} & 78.86 / 89.41 & 821 \\
 & UD-Naija-NSC & \textbf{94.56} / \textbf{92.71} & 94.56 / 92.71 & 68.66 / 78.96 & 759 \\
 & UD-Yoruba-YTB & 93.41 / 93.88 & \textbf{93.8} / \textbf{94.19} & 71.2 / 81.83 & 81 \\
   \midrule
persian & UD-Persian-Seraji & 96.15 / \textbf{96.85} & \textbf{95.95} / 96.69 & - / - & 4798 \\
  \midrule
phillipine & UD-Tagalog-TRG & 83.78 / 92.09 & \textbf{83.78} / \textbf{92.75} & 44.0 / 69.31 & 45 \\
  \midrule
sinotibetan & UD-Cantonese-HK & \textbf{89.64} / \textbf{86.82} & 89.64 / 86.82 & 70.15 / 77.76 & 521 \\
 & UD-Chinese-CFL & \textbf{88.65} / \textbf{86.96} & 88.65 / 86.96 & 74.65 / 79.91 & 361 \\
 & UD-Chinese-GSD & 90.83 / 90.54 & \textbf{90.9} / \textbf{90.56} & 76.81 / 84.35 & 3998 \\
 & UD-Vietnamese-VTB & \textbf{90.1} / \textbf{88.84} & 90.1 / 88.84 & 70.71 / 79.01 & 2401 \\
   \midrule
semitic & UD-Akkadian-PISANDUB & 79.21 / 78.65 & 79.21 / 78.65 & \textbf{84.0} / \textbf{84.19} & 81 \\
 & UD-Amharic-ATT & 87.24 / \textbf{91.13} & \textbf{86.58} / 90.91 & 76.0 / 88.16 & 860 \\
 & UD-Arabic-PADT & 91.77 / \textbf{95.44} & \textbf{91.52} / 95.36 & 77.03 / 92.03 & 6132 \\
 & UD-Arabic-PUD & 77.63 / \textbf{89.06} & \textbf{77.89} / 89.0 & 63.81 / 86.29 & 801 \\
 & UD-Hebrew-HTB & \textbf{94.33} / \textbf{95.81} & 94.03 / 95.65 & 81.59 / 91.84 & 4973 \\

\bottomrule
 \end{tabular}
 }
 \caption{Comprehensive results }
   \label{tab:result1}
\end{center}
 \end{table*}

\begin{table*}
\small
 \begin{center}\resizebox{\textwidth}{!}{
\begin{tabular}{c|l|c|c|c|c}
 \textbf{Cluster} & \textbf{Target} & \textbf{\textsc{\small{multi-source}}} & \textbf{\textsc{\small{multi-source}}}  &\textbf{\cite{sigmorphon2019}} & \textbf{\# Training} \\
  & & \textbf{\textsc{\small{+ Self-Attention}}} & & &  \textbf{Sentences} \\
  & & \textbf{Accuracy / F1} & \textbf{Accuracy / F1}& \textbf{Accuracy / F1} & \\
 \toprule
  turkic & UD-Turkish-IMST & \textbf{85.68} / \textbf{90.64} & 85.02 / 90.43 & 62.04 / 85.33 & 4509 \\
 & UD-Turkish-PUD & \textbf{79.78} / \textbf{90.88} & 79.33 / 90.54 & 66.92 / 88.05 & 801 \\
 \midrule
romance & UD-Catalan-AnCora & textbf{96.68} / \textbf{98.26} & 96.63 / 98.24 & 85.77 / 95.7 & 13343 \\
 & UD-French-GSD & 96.19 / \textbf{97.51} & \textbf{95.76} / 97.32 & 84.44 / 94.81 & 13074 \\
 & UD-French-ParTUT & 93.04 / 96.05 & \textbf{93.04} / \textbf{96.12} & 81.32 / 92.08 & 817 \\
 & UD-French-Sequoia & \textbf{95.08} / 96.95 & 94.96 / \textbf{96.96} & 82.64 / 93.42 & 2480 \\
 & UD-French-Spoken & \textbf{96.05} / \textbf{96.08} & 96.05 / 96.08 & 94.57 / 94.85 & 2229 \\
 & UD-Galician-CTG & 96.65 / 96.31 & \textbf{96.66} / \textbf{96.32} & 87.23 / 91.81 & 3195 \\
 & UD-Galician-TreeGal & \textbf{89.69} / 93.2 & 89.3 / \textbf{93.25} & 76.85 / 90.05 & 801 \\
 & UD-Italian-ISDT & 95.91 / 97.24 & \textbf{95.96} / \textbf{97.27} & 83.62 / 94.34 & 11334 \\
 & UD-Italian-ParTUT & \textbf{95.0} / 96.39 & 94.87 / 96.39 & 84.03 / 93.42 & 1673 \\
 & UD-Italian-PoSTWITA & \textbf{92.13} / \textbf{93.13} & 92.03 / 93.02 & 70.23 / 88.18 & 5371 \\
 & UD-Italian-PUD & \textbf{87.55} / 92.4 & 87.38 / \textbf{92.46} & 80.89 / \textbf{92.66} & 801 \\
 & UD-Portuguese-Bosque & \textbf{92.28} / \textbf{95.57} & 92.06 / 95.5 & 63.14 / 86.12 & 7493 \\
 & UD-Portuguese-GSD & \textbf{97.33} / \textbf{97.54} & 97.33 / 97.54 & - / - & 9663 \\
 & UD-Romanian-Nonstandard & \textbf{91.13} / \textbf{95.33} & 91.07 / 95.29 & 74.31 / 91.5 & 8056 \\
 & UD-Romanian-RRT & 94.67 / 96.58 & \textbf{94.82} / \textbf{96.63} & 81.45 / 93.96 & 7620 \\
 & UD-Spanish-AnCora & \textbf{96.97} / \textbf{98.25} & 96.86 / 98.22 & 84.27 / 95.3 & 14145 \\
 & UD-Spanish-GSD & 94.05 / 97.08 & \textbf{94.07} / \textbf{97.1} & - / - & 12811 \\
  \midrule
 slavic & UD-Belarusian-HSE & \textbf{79.63} / \textbf{85.37} & 77.28 / 84.11 & 54.99 / 79.07 & 315 \\
 & UD-Bulgarian-BTB & \textbf{94.22} / \textbf{96.44} & 93.99 / 96.37 & 79.75 / 93.91 & 8911 \\
 & UD-Buryat-BDT & \textbf{78.85} / \textbf{81.24} & 75.96 / 78.66 & 63.26 / 78.53 & 742 \\
 & UD-Old-Church-Slavonic-PROIEL & \textbf{87.22} / \textbf{94.13} & 86.94 / 94.03 & 82.86 / 90.34 & 5070 \\
 & UD-Russian-GSD & \textbf{84.26} / \textbf{91.91} & 83.25 / 91.55 & 64.42 / 88.77 & 4025 \\
 & UD-Russian-PUD & 76.77 / \textbf{87.55} & \textbf{77.25} / 87.49 & 63.15 / 85.52 & 801 \\
 & UD-Russian-SynTagRus & 91.65 / 95.96 & \textbf{92.74} / \textbf{96.5} & 73.9 / 92.84 & 49512 \\
 & UD-Russian-Taiga & 74.14 / 80.23 & \textbf{75.24} / \textbf{81.25} & 52.99 / 78.71 & 1412 \\
 & UD-Ukrainian-IU & \textbf{86.02} / \textbf{92.41} & 85.33 / 92.2 & 63.36 / 87.01 & 5441 \\
 & UD-Upper-Sorbian-UFAL & \textbf{74.04} / \textbf{82.45} & 70.12 / 81.21 & 55.66 / 78.3 & 517 \\
  \midrule
 ugric & UD-Estonian-EDT & 87.71 / 94.58 & \textbf{88.47} / \textbf{94.93} & 74.56 / 91.71 & 24579 \\
 & UD-Finnish-FTB & 83.24 / 90.38 & \textbf{83.63} / \textbf{90.7} & 73.16 / 89.51 & 14979 \\
 & UD-Finnish-PUD & 77.05 / 86.33 & \textbf{77.49} / \textbf{86.77} & 71.65 / \textbf{88.87} & 801 \\
 & UD-Hungarian-Szeged & \textbf{80.57} / \textbf{90.88} & 79.16 / 90.13 & 63.72 / 87.29 & 1441 \\
 & UD-North-Sami-Giella & \textbf{84.35} / \textbf{88.8} & 83.78 / 88.65 & 67.04 / 85.6 & 2498 \\
 & UD-Norwegian-Bokmaal & \textbf{94.97} / \textbf{96.68} & 94.58 / 96.51 & 81.44 / 93.19 & 16037 \\
 & UD-Swedish-Talbanken & \textbf{93.94} / \textbf{96.01} & 93.64 / 95.9 & - / - & 4821 \\
 & UD-Finnish-TDT & \textbf{86.51} / \textbf{92.63} & 85.55 / 92.2 & 75.13 / 90.92 & 12109 \\
  \midrule
westslavic & UD-Croatian-SET & \textbf{87.23} / \textbf{94.04} & 86.88 / 93.91 & 72.71 / 90.99 & 7112 \\
 & UD-Czech-CAC & 90.66 / 96.72 & \textbf{91.38} / \textbf{96.99} & 77.15 / 93.92 & 19768 \\
 & UD-Czech-CLTT & \textbf{91.29} / 96.15 & 91.07 / \textbf{96.22} & 73.92 / 92.37 & 901 \\
 & UD-Czech-FicTree & 90.05 / 95.42 & \textbf{90.0} / \textbf{95.49} & 68.28 / 90.37 & 10209 \\
 & UD-Czech-PDT & \textbf{89.78} / \textbf{96.37} & 54.13 / 73.56 & 76.69 / 94.28 & 70331 \\
 & UD-Czech-PUD & 75.65 / 88.19 & \textbf{77.72} / \textbf{89.37} & 59.54 / 85.5 & 801 \\
 & UD-Polish-LFG & 87.76 / \textbf{93.7} & \textbf{87.81} / 93.65 & - / - & 13797 \\
 & UD-Polish-SZ & \textbf{82.27} / \textbf{91.38} & 81.01 / 90.88 & 65.58 / 88.29 & 6582 \\
 & UD-Serbian-SET & \textbf{91.89} / \textbf{95.46} & 91.35 / 95.29 & 75.73 / 91.19 & 3113 \\
 & UD-Slovak-SNK & \textbf{85.59} / \textbf{93.12} & 84.99 / 92.83 & 64.24 / 88.16 & 8484 \\
 & UD-Slovenian-SSJ & \textbf{89.05} / \textbf{94.03} & 87.92 / 93.55 & 73.73 / 89.95 & 6401 \\
 & UD-Slovenian-SST & 85.13 / \textbf{90.16} & \textbf{85.51} / 90.02 & 73.4 / 84.74 & 2551 \\
\bottomrule
 \end{tabular}
 }
 \caption{Comprehensive results }
   \label{tab:result2}
\end{center}
 \end{table*}

%% file: Amethod.tex
\section{Language Clusters}
\label{lang}
\begin{figure*}%
    \centering
    \includegraphics[width=1.0\textwidth]{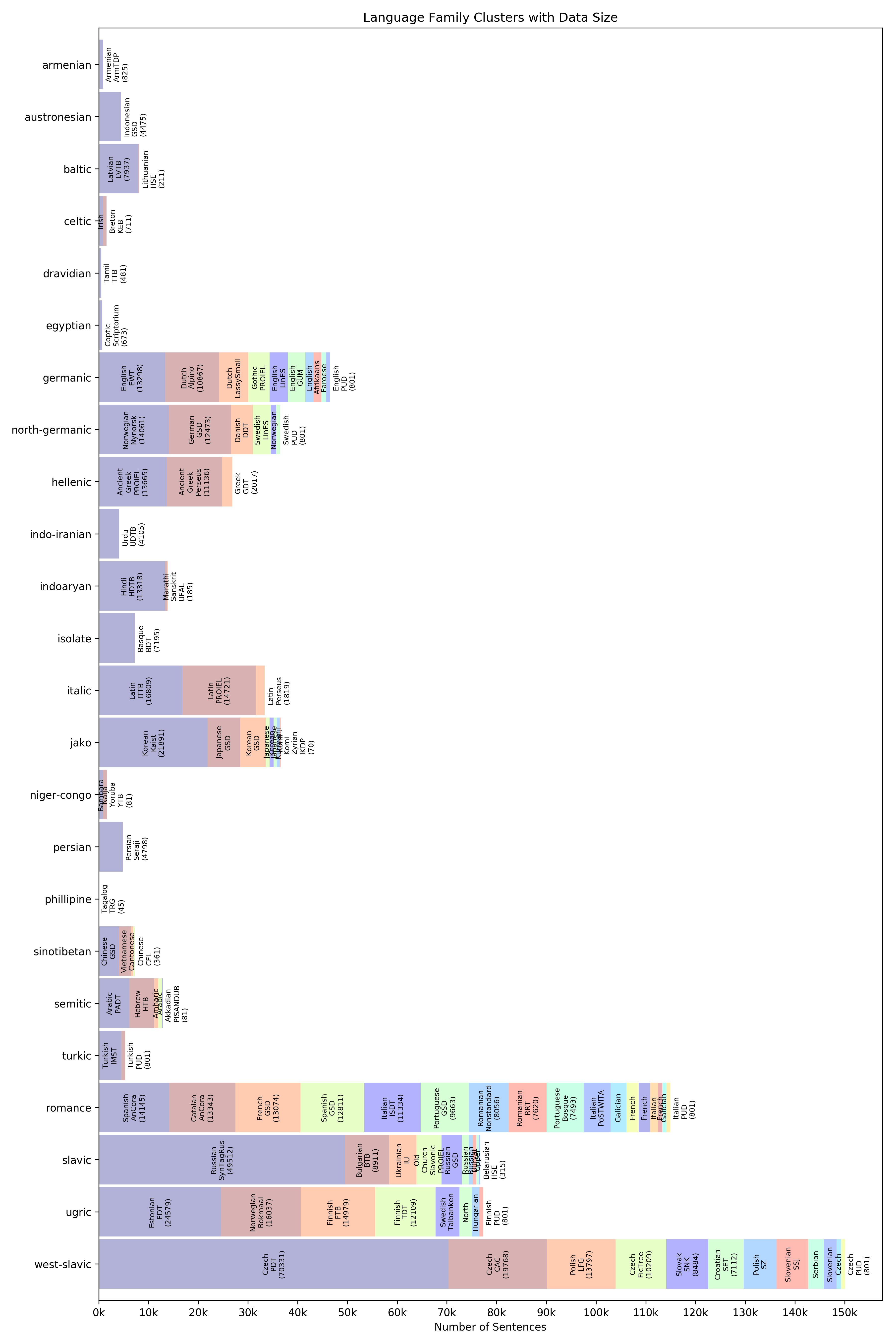} 
  \caption{Language family clusters, by number of sentences per dataset.}%
\label{fig:clusters}%
\end{figure*}

We train one model per language cluster for the multi-lingual transfer learning. Each language cluster was constructed based on the typological and orthographic similarity of the languages therein. Table \ref{tab:result1}, \ref{tab:result2} show details of the language clusters. Figure \ref{fig:clusters} shows clusters graphically by relative size per dataset. 

